\documentclass[10pt,twocolumn,letterpaper]{article}

\usepackage{cvpr}              
\usepackage{multirow}
\usepackage{array}
\usepackage{xcolor}
\usepackage{colortbl}
\usepackage{amssymb}
\usepackage{overpic}
\newcolumntype{C}[1]{>{\centering\arraybackslash}m{#1}}
\definecolor{cvprblue}{rgb}{0.21,0.49,0.74}
\usepackage[pagebackref,breaklinks,colorlinks,allcolors=cvprblue]{hyperref}

\def\paperID{2266} 
\def\confName{CVPR}
\def\confYear{2026}

\title{Event-based Visual Deformation Measurement}

\newcommand\blfootnote[1]{%
  \begingroup
  \renewcommand\thefootnote{}\footnote{#1}%
  \addtocounter{footnote}{-1}%
  \endgroup
}

\author{
    Yuliang Wu\textsuperscript{1}, Wei Zhai\textsuperscript{1,$\dagger$}, Yuxin Cui\textsuperscript{1}, Tiesong Zhao\textsuperscript{2}, Yang Cao\textsuperscript{1}, Zheng-Jun Zha\textsuperscript{1} \\[0.2cm]
    \textsuperscript{1}MoE Key Laboratory of Brain-inspired Intelligent Perception and Cognition,\\ University of Science and Technology of China \qquad
    \textsuperscript{2}Fuzhou University \\
    {\tt\small \{tronliang@mail., wzhai056@, yuxincui@mail.\}ustc.edu.cn}\\
    {\tt\small t.zhao@fzu.edu.cn}\qquad
    {\tt\small \{forrest@, zhazj@\}ustc.edu.cn}
}

\begin{document}
\maketitle
\blfootnote{$^\dagger$Corresponding author.}

\begin{abstract}

Visual Deformation Measurement (VDM) aims to recover dense deformation fields by tracking surface motion from camera observations. 
Traditional image-based methods rely on minimal inter-frame motion to constrain the correspondence search space, which limits their applicability to highly dynamic scenes or necessitates high-speed cameras at the cost of prohibitive storage and computational overhead.
We propose an event-frame fusion framework that exploits events for temporally dense motion cues and frames for spatially dense precise estimation.
Revisiting the solid elastic modeling prior, we propose an Affine Invariant Simplicial (AIS) framework. It partitions the deformation field into linearized sub-regions with low-parametric representation, effectively mitigating motion ambiguities arising from sparse and noisy events.
To speed up parameter searching and reduce error accumulation, a neighborhood-greedy optimization strategy is introduced, enabling well-converged sub-regions to guide their poorly-converged neighbors, effectively suppress local error accumulation in long-term dense tracking.
To evaluate the proposed method, a benchmark dataset with temporally aligned event streams and frames is established, encompassing over 120 sequences spanning diverse deformation scenarios. 
Experimental results show that our method outperforms the state-of-the-art baseline by 1.6× in survival rate. Remarkably, it achieves this using only 18.9\% of the data storage and processing resources of high-speed video methods.

\end{abstract}    
\section{Introduction}
\label{sec:intro}
\begin{figure}[tp]
  \centering
  \includegraphics[width=1\linewidth]{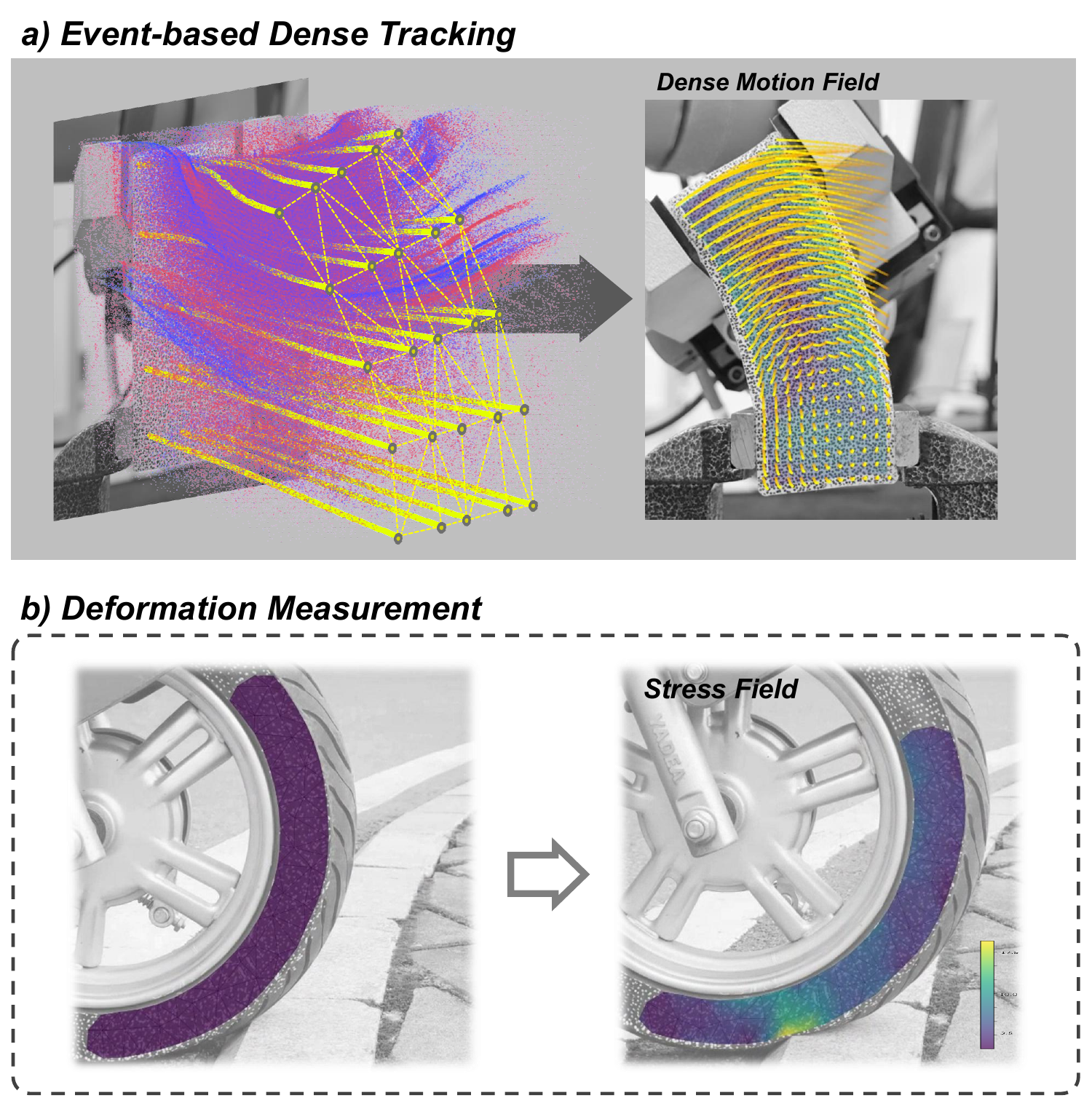}
  \caption{ Overview of Event-based Deformation Measurement. (a) Our approach regresses deformation fields by leveraging affine-invariant spatiotemporal trajectory flows extracted from both events and images. (b) Measuring tire deformation while rolling over a step obstacle. Demonstrating robust and accurate measurement capabilities of our event-based VDM approach for objects experiencing rapid self-motion.}
  \label{fig:mainfig}
\end{figure}
Deformation is ubiquitous in the real world, ranging from subtle material strain to large-scale structural changes. Visual Deformation Measurement (VDM) recovers deformation fields by accurately tracking surface motion from camera observations, serving as an efficient non-contact technique for structural monitoring \cite{liu2024experimental,ri2024drone}, mechanics analysis \cite{luo2024strain,katz2020new,cao2021displacement}, and biomechanics research \cite{yoon2021digital}.

As a dense tracking vision problem, the goal of VDM is to estimate the surface motion of a target at the current moment relative to its initial undeformed state. 
However, unlike rigid motion with only 6-DoF, deformable surfaces have significantly higher degrees of freedom where surface points can move with considerable independence. This poses fundamental challenges to image-based methods: (i) an intractably large correspondence search space through  (ii) texture similarity and severe appearance/geometric changes from deformation that impede reliable feature matching.
To mitigate these challenges, existing image-based deformation measurement methods rely on costly high-speed cameras to capture videos with minimal inter-frame motion, thereby constraining the correspondence search space. However, this strategy incurs prohibitive storage and computational costs from processing massive redundant frames, limiting the practicality and scalability of such systems.

In this paper, we address these challenges by introducing an event-frame hybrid system that exploits the complementary nature of two vision modalities. Events, being temporally dense yet spatially sparse, capture motion dynamics efficiently without the massive redundancy of high-speed videos. Conversely, conventional frames offer spatially dense, low-noise observations to ensure precise dense field estimation. This synergy enables accurate deformation tracking without the prohibitive costs of traditional high-speed imaging.
However, employing the hybrid system for deformation measurement presents numerous challenges.
Firstly, the sparse spatiotemporal sampling inherent to event cameras, coupled with their high noise levels, introduces ambiguities in the estimation of dense field motion information.

Secondly, long-term tracking suffers from error accumulation, which is significantly amplified in dense tracking due to the large number of descriptors at each time step. The accumulated local errors can potentially lead to global failure.

In this paper, we propose an Affine Invariant Simplicial (AIS) framework that leverages a simplicial structure to locally linearize the deformation field, reduce the number of motion field descriptors and mitigate motion estimation ambiguities.
Meanwhile, we introduce a neighborhood-greedy optimization strategy that corrects mismatched sub-regions using their well-converged neighbors, thereby reducing error accumulation in dense tracking and achieving long term global optimality.

In the proposed AIS framework, the nonlinear deformation field of the object surface is systematically divided into multiple sub-regions. Within each sub-region, deformation parameters are linearized using simplicial motion parameterization, effectively reducing their dimensionality.
Then, leveraging event contrast maximization and image cross-correlation as optimization objectives, a coarse-to-fine scheme is adopted to hierarchically decouple and sequentially solve for: rigid self-motion and fine-grained deformations across multi-level mesh subdivisions. 
The AIS framework hits two birds with one stone: it reduces the difficulty of solving the high-dimensional deformation field and addresses the ambiguity in motion information estimation from event data.
Building upon the framework, we employ the neighborhood-greedy optimization strategy. We design a global convergence criterion to distinguish poorly-converged subregions that significantly deviate from mean convergence states. Then, Leveraging continuity priors, the optimization of this subregions are guided by their well-converged counterparts, significantly reducing long-term error accumulation in dense tracking.

To establish a comprehensive evaluation benchmark for our method and a series of baselines, we captured a dataset comprising over 120 real-world sequences spanning various deformation modes and magnitudes, with temporally aligned event and frame data. Ground truth deformation fields derived from high-speed videos are provided to facilitate quantitative evaluation and advance research.

In summary, our contributions are as follows.
\begin{enumerate}
  \item A hybrid VDM system that leverages the high temporal resolution of event cameras to capture rapid texture motion, enabling accurate measurement of large-scale deformations and those involving significant self-motion.

  \item An affine invariant simplicial framework is proposed for robust deformation measurement from events and frames. This formulation retains the expressive power for complex deformation fields while mitigating the motion estimation ambiguity and noise sensitivity inherent to spacial-temporal sparse event data.

  \item A neighborhood-greedy optimization is proposed to exploits continuity priors to enable well-converged sub-regions to guide their poorly-converged neighboring regions, mitigating suboptimal error accumulation in long term dense deformation field tracking.
  \item Extensive experiments demonstrate the competitiveness of the proposed approach, achieving a survival rate of 65.7\% on samples with 100+ pixel displacements (1.6× that of SOTA method), while maintaining superior EPE accuracy with only 18.9\% of the data storage and processing cost of traditional high-speed video methods.

\end{enumerate}

\section{Related Work}
\label{sec:related work}

\subsection{Visual Deformation Measurement}

As an efficient non-contact deformation sensing approach, visual deformation measurement (VDM) has been widely applied in scientific research \cite{palanca2016use,genovese2013improved,zhou2016using,krehbiel2010digital,huang2010high,verbruggen2015altered,yoon2021digital} and engineering fields \cite{katz2020new,cao2021displacement}.
Image-based VDM algorithms use correlation criteria to match intensity values between regions before and after deformation,
\begin{equation}
\label{eq:dic}
C=\sum_{i=1}^{N}\left(a f\left(x_{i}, y_{i}\right)+b-g\left(x_{i}^{\prime}, y_{i}^{\prime}\right)\right)^{2}.
\end{equation}
Where $(x_{i},y_{i})\in R$ and $(x_{i}^{\prime}, y_{i}^{\prime})\in R'$ denote corresponding point pairs in the original and deformed regions, respectively. Here, $a$ represents a scale factor and $b$ represents an intensity offset. The task of the VDM is to minimize the
coefficient $C$ in Eq. \ref{eq:dic} to detect the best matching.
However, unlike rigid body motion tracking in everyday scenarios that only requires estimating a limited number of parameters, VDM is fundamentally a dense tracking task that demands high accuracy, involves complex variations, and requires optimization over a large parameter space, making it a complex high-dimensional optimization problem. To reduce the complexity of subset matching, VDM methods adopt the spatiotemporal continuity prior of motion and search for matching subsets in the vicinity of the original subset positions \cite{mguil2024various,he2024review}.

This limitation prevents them from effectively handling large deformations and large displacement scenarios with rapid object self-motion \cite{boukhtache2023lightweight,wang2023dic,yang2022deep,yang2023r3}, or necessitates the use of costly high-speed cameras, which introduces a significant data storage burden, greatly restricting the potential applications of the VDM method.

\subsection{Event-based Vision}
Different from traditional frame-based camera, the neuromorphic event camera abandons the fixed-interval sampling approach. Instead, it uses trigger-based sampling to passively capture dynamic information from the scene \cite{gehrig2024low,gallego2020event,brandli2014240,serrano2013128,wan2025emotive,liao2024ef,han2024event,wan2024event}. Specifically, the i-th event $e_{i}:=\left(x_{i}, y_{i}, t_{i}, p_{i}\right)$ is triggered at time $t_{i}$ whenever the log-scale scene brightness $\mathbf{I}(t)=\log \left(I_{x_{i},y_{i}}\left(t\right)\right)$ change exceeds threshold $c$, i.e,
{\small
\begin{equation}
    p_{i}=\left\{\begin{array}{ll}
    1, & \text { if } \mathbf{I}(t_i)-\mathbf{I}(t_{i}-\Delta t_{i}) \geq c, \\
    -1, & \text { if } \mathbf{I}(t_i)-\mathbf{I}(t_{i}-\Delta t_{i}) \leq-c,
    \end{array}\right.
\end{equation}
}
Here, $\Delta t_{i}$ indicates the time since the last event at $(x_{i},y_{i})$, and $p_{i} \in\{-1,+1\}$ denotes the event polarity.
This sampling strategy provides event cameras with high temporal resolution \cite{tulyakov2021time,tulyakov2022time,chen2024event}, high dynamic range \cite{yang2023learning,mostafavi2021learning,zou2024eventhdr,wu2024event}, and low storage redundancy \cite{hamann2024low}, making them ideal for  continuous motion observation in non-rigid shape estimation and reconstruction tasks \cite{millerdurai20243d,xue2024event,xue2022event,nehvi2021differentiable}.

\subsection{Contrast Maximization}
Currently, there are various model-based \cite{gallego2018unifying,stoffregen2019event,gehrig2020eklt} and learning-based  \cite{zhu2018ev,gehrig2021raft,gehrig2024dense,shiba2024secrets} approaches being capable of motion tracking from event streams at the pixel level. 
Regrettably, there is currently no event dataset suitable for VDM (Visual Deformation Modeling) tasks. Moreover, creating large-scale deformation datasets is costly and inefficient. In this work, the data-independent contrast maximization (CM) method is employed for motion estimation.
As a versatile method, CM works by maximizing the contrast of the image of warped events (IWE) to find the motion field that best fits the event set \cite{gallego2018unifying,stoffregen2019event,gehrig2024dense}. This allows it to be more robust than the fixed-template matching methods \cite{gehrig2020eklt}, especially for handling feature shape changes caused by deformations. 
However, due to spatiotemporal sparsity and high noise level in event data, pixel-level CM-based motion estimation suffers from significant ambiguity. Existing CM frameworks address this through low-parameter modeling for global motion (e.g., camera ego-motion or planar scenes\cite{gallego2018unifying}) or smoothness/regularization constraints for local neighborhoods \cite{hamann2024motion}. However, neither approach applies to VDM tasks: deformations lack global parametric models, while smoothness constraints suppress the framework's ability to capture deformation details.
Incorporating physical priors, such as solid elastic modeling, offers a promising solution to capture complex deformable behaviors
\cite{agudo2014good,malti2017elastic,agudo2015sequential,kairanda2022f}.
In this paper, a simplicial parameterization framework is proposed to address the dilemma. 
The surface region of the deformed object is divided into a set of continuous local sub-regions and simplify the deformation of each sub-region as a low-parameter description. This allows achieving low-ambiguity CM motion estimation while preserving the ability to describe the deformation.

\section{Methodology}
In this section, we present our approach for dense tracking of deformable objects using hybrid event-frame data.
Sec. 3.1 formulates the VDM problem and introduces the affine-invariant property that enables efficient field representation.
Sec. 3.2 describes how we decompose the deformation field into sparse anchor trajectories and associate events/images with these anchors through geometric operations.
Sec. 3.3 details our coarse-to-fine and neighborhood-greedy optimization strategy for efficient solving.
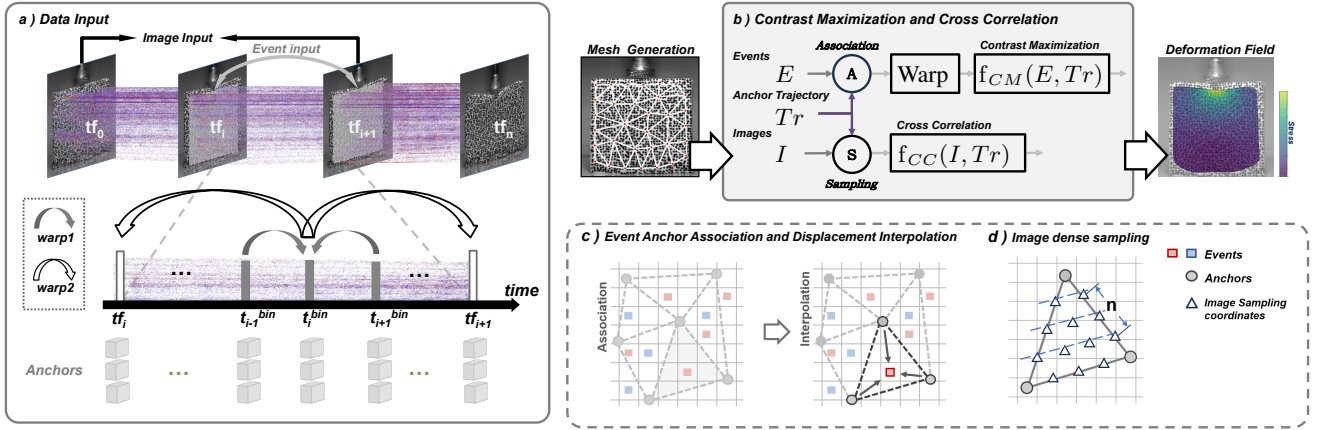
\begin{figure*}[tp]
  \centering
  \begin{overpic}[width=1.\linewidth]{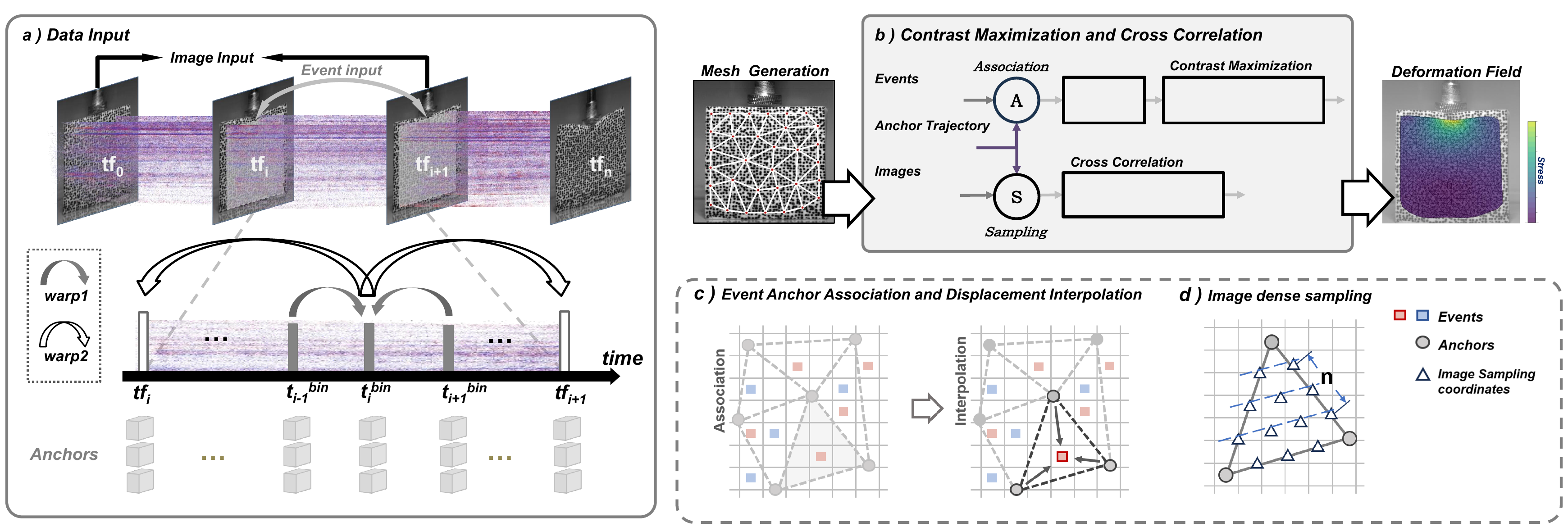}
  \put(59.0,26.6){\small{$E$}}
  \put(68.3,26.6){\small Warp}
  \put(74.5,26.6){\small{$\operatorname{f}_{CM}(E,Tr)$}}
  \put(59.0,23.35){\small{$Tr$}}
  \put(59.0,20.5){\small{$I$}}
  \put(68.35,20.5){\small{$\operatorname{f}_{CC}(I ,Tr)$}}
  \end{overpic}
  \caption{\textbf{Illustration of the key steps of our method.} (a) Data input and event warping strategy. 
(b) Overall pipeline of the proposed event-based VDM framework. 
(c) Illustration of event-to-subregion association through vector cross product with anchor trajectories and subsequent displacement interpolation. 
(d) Image intensity sampling scheme within each subregion.}
  \label{fig:method}
\end{figure*}
\subsection{Preliminaries}
\textbf{VDM problem.} The goal of VDM is to recover the surface mapping relationship before and after object deformation through visual methods, i.e., the deformation field $u$.
Under the Lagrangian perspective, we formulate the problem as follows.
At any time $t_i$, a material point initially located at position $ X \in S$ on the object surface 
$S$ deforms to its current position:
{\small
\begin{equation}
x\left(X, t_{i}\right)=X+u\left(X, t_{i}\right), \quad X \in S,
\end{equation}}
where $u(X,t)$ represents the displacement vector of the material point initially located at $X$ on surface $S$.
In traditional image-based tasks, this is defined as a simple mapping between two frames, whereas in event-based VDM, we need to construct a data structure to represent the deformation field in continuous spatiotemporal space.

\textbf{Affine-invariant property.} For any given triangular sub-region defined as:
$\sigma = \operatorname{conv}\left\{X_{1}, X_{2}, X_{3}\right\}$\footnote{The $conv \left\{ . \right\}$ operator denotes the convex hull}, the affine interpolation operator can be expressed as:
{\small
\begin{equation}
    I[f](X)=\sum_{i=1}^{3} \lambda_{i}(X) f\left(X_{i}\right),
    \quad \text{s.t.} \sum_{i=1}^{3} \lambda_{i}(X) = 1.
    \label{eq:invariant}
\end{equation}}

For any arbitrary affine function of the form
$f(X)=A X+b$, the interpolation operator satisfies the reproduction property $I[f]=f$. This fundamental property enables modeling of spatiotemporal deformation fields via low-parametric sparse triangle sub-regions.

\subsection{Dense Tracking with Simplicial Framework }
In this paper, we introduce the affine invariant simplicial framework (AIS), which represents the spatiotemporal deformation field as lightweight trajectories $\{\text{Tr}_j(t)\}$. 
Specifically, high-dimensional nonlinear deformation field $u$ is decomposed into $N$ triangular sub-regions $T_k, k=1, \ldots, SN$, 
where the deformed position within each triangle follows an affine transformation:
\begin{equation}
x(X)=X+u(X)=A_{k} X+b_{k}, \quad \forall X \in T_{k},
\end{equation}
$A_{k} \in \mathbb{R}^{2 \times 2}$ is the local deformation gradient matrix and $b_{k} \in \mathbb{R}^{2}$ is the translation vector.
Larger $SN$ improves nonlinear expressiveness but complicates optimization, while smaller $SN$ offers simplicity with reduced none-linear representation ability.
Stems from the eq. \ref{eq:invariant} property, the affine transformation of each sub-region can be losslessly described by the motion of its vertices, we optimize the trajectories of these anchor points $\{\text{Tr}_j(t)\}$ using events and images to obtain the continuous spatiotemporal field $u(t)$.

\textbf{Event anchor association and displacement interpolation.} During optimization process, the association between individual events and their corresponding trajectories is unknown, and efficiently establishing this association is a key challenge. Prior work \cite{hamann2024motion} employs knn search to find the top-k nearest anchors for each event, then computes displacements via averaging. However, per-event KNN search is time-consuming, and averaging-based interpolation introduce significant errors in deformation measurement.
In this work, we perform same-side test event association and using the affine invariant interpolation to address this issue, as shown in Fig. \ref{fig:method} (d.
Consider the spatiotemporal motion of an individual sub-region $\sigma(t) = conv\left\{\text{Tr}_{1}(t), \text{Tr}_{2}(t), \text{Tr}_{3}(t)\right\}$. 

The same-side Test association is performed as follows.
Given event coordinate $\mathbf{X}_{e}=\left[x_{i}, y_{i}\right]^{T}$ and triangle vertices at triggering time $\text{Tr}_{j}(t_{i})=\left[x^{t}_{j}, y^{t}_{j}\right]^{T}$
, we compute:
\begin{equation}
C_{i}=\left|\begin{array}{cc}
x^{t}_{(j \bmod 3)+1}-x^{t}_{j} & y^{t}_{(j \bmod 3)+1}-y^{t}_{j} \\
x_{i}-x^{t}_{j} & y_{i}-y^{t}_{j}
\end{array}\right|,
\end{equation}
where $j \in\{1,2,3\}$,
and the event is inside if:
\begin{equation}
\mathbf{X}_{e} \in \sigma(t) \Longleftrightarrow \operatorname{sign}\left(C_{1}\right)=\operatorname{sign}\left(C_{2}\right)=\operatorname{sign}\left(C_{3}\right)
\end{equation}
Once associated with a subregion, the barycentric interpolation weights $\lambda_i$ are computed by solving:
\begin{equation}
\mathbf{X}{e} = \sum{k=1}^{3} \lambda_k \text{Tr}_{j}(t_{i}), \quad \text{s.t.} \sum{k=1}^{3} \lambda_k = 1,
\end{equation}
The solving process are provided in appendix.
For any selected timestamp $t_{ref}$, the event can be warped according to $\mathbf{X}_{e}^{\prime} = \left[x_{i}^{\prime}, y_{i}^{\prime}\right]^{T} = \sum_{k=1}^{3} \lambda_{k} \text{Tr}_{j}(t_{ref})$ to form the image of warped events (IWE).
Following the formulation in \cite{hagenaars2021self,paredes2023taming}, the IWE is constructed as:
\begin{equation}
\begin{array}{c}
T_{p}\left(\boldsymbol{x}, t_{\text {ref }}\right) =\frac{\sum_{j} \kappa\left(x-x_{j}^{\prime}\right) \kappa\left(y-y_{j}^{\prime}\right) w(t_{j}) }{\sum_{j} \kappa\left(x-x_{j}^{\prime}\right) \kappa\left(y-y_{j}^{\prime}\right)+\epsilon},\\
w\left(t_{j}\right)=1-\frac{\left|t_{r e f}-t_{j}\right|}{\max _{i}\left|t_{r e f}-t_{i}\right|},\\
\kappa(a) =\max (0,1-|a|),\\
j=\left\{i \mid p_{i}=p\right\}, \quad p^{\prime} \in\{+1,-1\}, \quad \epsilon \approx 0,
\end{array}
\end{equation}
And the contrast maximization objective is formulated as:
\begin{equation}
\operatorname{f_{CM}}\left(t_{\text {ref }}\right)=\frac{\sum_{\boldsymbol{x}} T_{+1}\left(\boldsymbol{x},t_{\text {ref}}\right)^{2}+T_{-1}\left(\boldsymbol{x},t_{\text {ref}}\right)^{2}}{\sum_{\boldsymbol{x}}\left[n\left(\boldsymbol{x}^{\prime}\right)>0\right]+\epsilon},
\end{equation}
where $n(\boldsymbol{x}^{\prime})$ denotes a per-pixel event count of the IWE.
To recover anchor motion trajectories, we warp the event stream to a set of reference timestamps 
${t_{ref}}$ and jointly optimize the motion parameters $\text{Tr}_{j}(t_{ref})$ by minimizing the contrast maximization objective.

\textbf{Image intensity sampling from anchors.} Due to the loss of precise grayscale information in sparse events, frames are introduced to refine the deformation field.
Specifically, $S_1$ and $S_2$ denote the sets of intensity samples at corresponding coordinates across two different frames $I_{1},I_{2}$. The coordinates are obtained through uniform barycentric sampling within sub-region $\sigma = conv\left\{\text{Tr}_{1}, \text{Tr}_{2}, \text{Tr}_{3}\right\}$ at frame capturing time:
{\small
\begin{equation}
 \mathcal{B} = \left\{\left(\frac{i}{n}, \frac{j}{n}, 1-\frac{i+j}{n}\right) \mid i,j \in \mathbb{N}, i+j \leq n\right\},
\end{equation}}
where $n$ is the number of samples per sub-region edge, and the sampling coordinates are given by 
$u\operatorname{Tr}_{1}+v \operatorname{Tr}_{2}+w \operatorname{Tr}_{3} \text { for each }\left(u,v,w\right) \in \mathcal{B}.$
Then the intensity at each sampling coordinate are obtained from image pixels via bilinear interpolation, and zero-mean normalized cross-correlation is employed to compute the cross-correlation between the intensity samples:
\begin{equation}
\operatorname{f_{CC}}\left(S_{1}, S_{2}\right)=\frac{\operatorname{Cov}\left(S_{1}, S_{2}\right)}{\sigma_{S_{1}} \cdot \sigma_{S_{2}}}.
\end{equation}
In practice, we optimize the $\operatorname{f_{CC}}$ between the current frame $I_{i}$ and both the previous frame $I_{i-1}$ and the initial frame $I_{0}$.

\subsection{Steps of the Optimization}
\begin{figure}[tp]
  \centering
  \includegraphics[width=1.\linewidth]{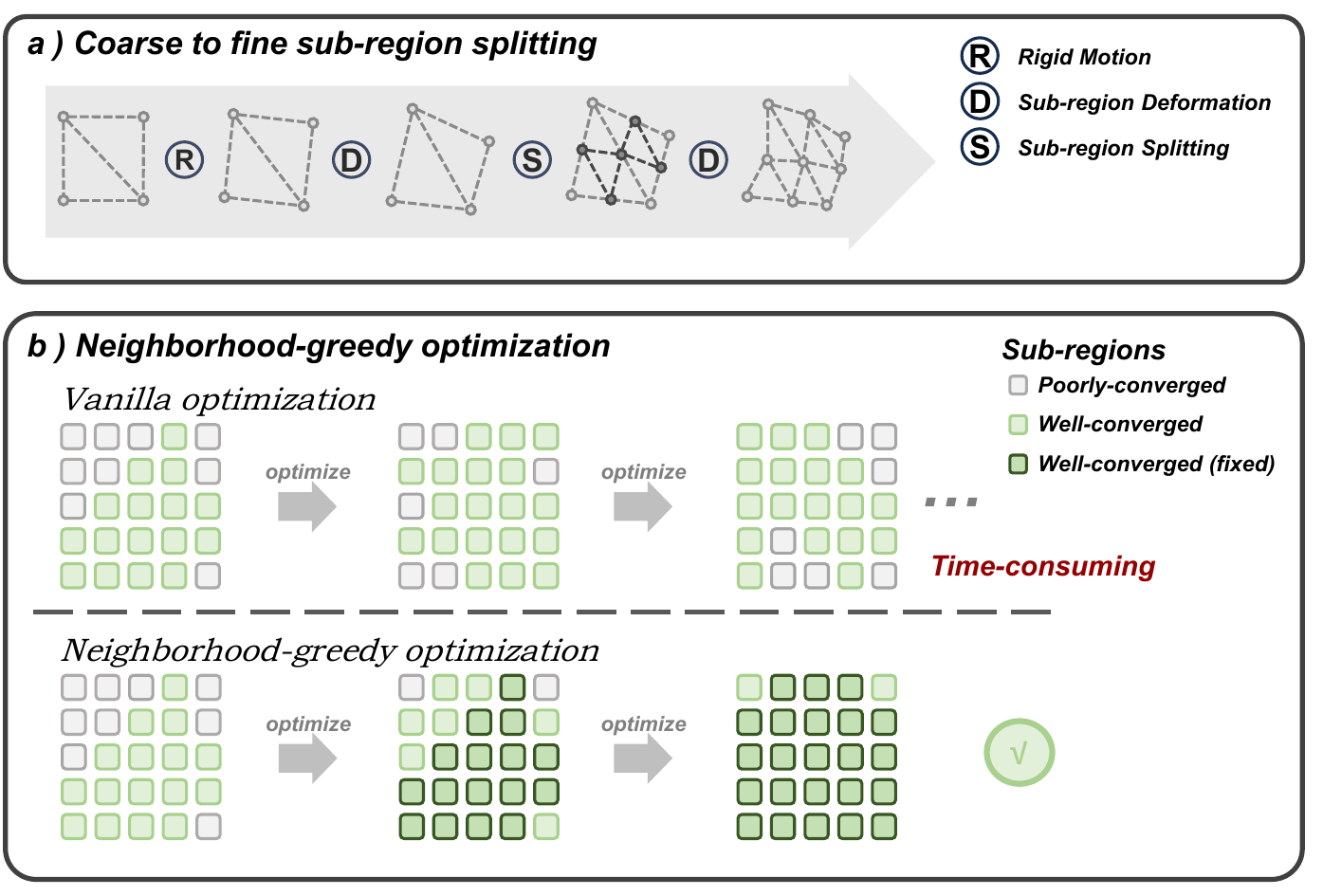}
  \caption{The optimization steps. (a) Hierarchical subregion optimization: we first estimate rigid motion parameters of the object, then progressively split and optimize subregions from coarse to fine. (b) Neighborhood-greedy optimization: well-optimized subregions serve as anchors to guide neighboring regions toward better convergence, achieving faster convergence to the global optimum compared to direct high dimensional optimization.}
  \label{fig:optsteps}
\end{figure}
The optimization pipeline iteratively estimates motion within each time window $tw$ in a forward manner.
In practice, the events $\mathcal{E}_{k,k+1}$ between consecutive frames $\{f_{k},f_{k+1}\}$ are partitioned into $M$ non-overlapping bins with equal numbers of events. 
Under the assumption that motion within each bin is linear, we optimize trajectories at $M+1$ timestamps.
\begin{equation}
\{\text{Tr}_j(t^{bin}_i)\}, i = 0, \ldots, M.
\end{equation}
Each event undergoes two warping operations to generate IWE for contrast maximization calculation, as shown in Fig.\ref{fig:method} a): \textbf{Warp 1}: For each time stamps $t^{bin}_i$, events from neighboring bins are warped to produce an $M$-channel IWE sequence, capturing short-term motion. \textbf{Warp 2}: All events within $tw$ are warped to $t_{f_i}$ and $t_{f_{i+1}}$, generating a 2-channel IWE to ensure global motion continuity.
The optimization pipeline follows a coarse-to-fine paradigm, as shown in Fig. \ref{fig:optsteps} a). It first estimates the rigid motion parameters of the target region, then performs initial deformation estimation on coarse subregions, and finally refines the deformation field through iterative splitting of these subregions.
The subdivision process is as follows:
{\small
\begin{equation}
\begin{aligned}
&\quad \quad \quad \quad\quad\text{conv}\{ \mathbf{Tr}_1,\mathbf{Tr}_2,\mathbf{Tr}_3\}\rightarrow \\ 
&\bigcup_{i=1}^{3} \text{conv}\{\mathbf{Tr}_i, \mathbf{M}_i, \mathbf{M}_{i-1}\} \cup \text{conv}\{\mathbf{M}_1, \mathbf{M}_2, \mathbf{M}_3\}, \\
&\quad \quad \quad M_i = \frac{\mathbf{Tr}_i + \mathbf{Tr}_{(i \bmod 3)+1}}{2}, \quad i = 1,2,3 .
\end{aligned}
\end{equation}}
The optimization objective is:
\begin{equation}
\operatorname{f}_{total} = \lambda_{1} \operatorname{f}_{CM}^{Warp1} + \lambda_{1} \operatorname{f}_{CM}^{Warp2} +  \lambda_{2} \operatorname{f}_{CC}, 
\end{equation}
where $\lambda_{i}$ are iteration-dependent coefficients.
This strategy achieves large-scale displacement estimation with low-parameters (rigid motion and coarse subregions), providing reliable initialization for high-parameter, fine-scale deformation estimation.

\textbf{Neighborhood-greedy optimization strategy.} While this pipeline achieves satisfactory convergence for most sub-regions, long-term tracking involves multiple iterations where errors from a small fraction of unmatched sub-regions can accumulate and lead to tracking failure. 
Therefore, it is crucial to pursue a global optimum at each time step to ensure robust and accurate tracking performance. 
However, jointly optimizing all sub-regions poses significant computational challenges due to the high-dimensional nature of the deformation parameters, making direct global optimization prohibitively time-consuming. To address this issue, we propose a neighborhood-greedy optimization strategy. The key idea is to leverage the continuity prior of deformation fields, using well-converged sub-regions to guide the optimization of their  poorly-converged neighbors, and greedily approaching the global optimum. First, the convergence quality of each subregion is assessed by computing the proportion of sampled pixels whose squared error (SE) deviates from the mean squared error (MSE) within the subregion $\Omega_{j}$:

{\small
\begin{equation}
\begin{aligned}
P_{j} & =\frac{1}{N} \sum_{i=1}^{N} 1(\mathrm{SE}(i)>k \cdot \mathrm{MSE}) \\
\quad \quad\operatorname{converged}\left(\Omega_{j}\right) & =\left\{\begin{array}{ll}
\text { False }, & \text { if } P_{j}>\tau \\
\text { True }, & \text { otherwise }
\end{array}\right.
\end{aligned}
\end{equation}}
where $k$ and $\tau$ are hyperparameters, set to 3 and 0.5 in our experiments. Then, anchor points associated with well-converged sub-regions are fixed, and strain continuity constraints $\operatorname{f}_{S}$ are added to the objective function for further optimization:
$\operatorname{f}_{S}= \frac{1}{|E_T|} \sum_{(i,j) \in E_T} \|\mathbf{S}_i - \mathbf{S}_j\|_F^2, $
where \(\mathbf{S}_i \in \mathbb{R}\) is the von-mises strain at anchor \(i\) (detailed computation in supplementary), and \(E_T\) denotes the set of edges of the sub-region.
As shown in Fig. \ref{fig:optsteps} b), after each optimization, the convergence status is re-evaluated and fixed anchors are greedily accumulated without release, until the global optimum is reached.
The following experiments \ref{tab:greedy ablation} show that the strategy both improves the long-term dense tracking survival rate and significantly reduces the convergence time of the optimization.

\section{Experiments}
\subsection{Event-based VDM Benchmark}

\begin{figure}[t]
  \centering
  \includegraphics[width=1.0\linewidth]{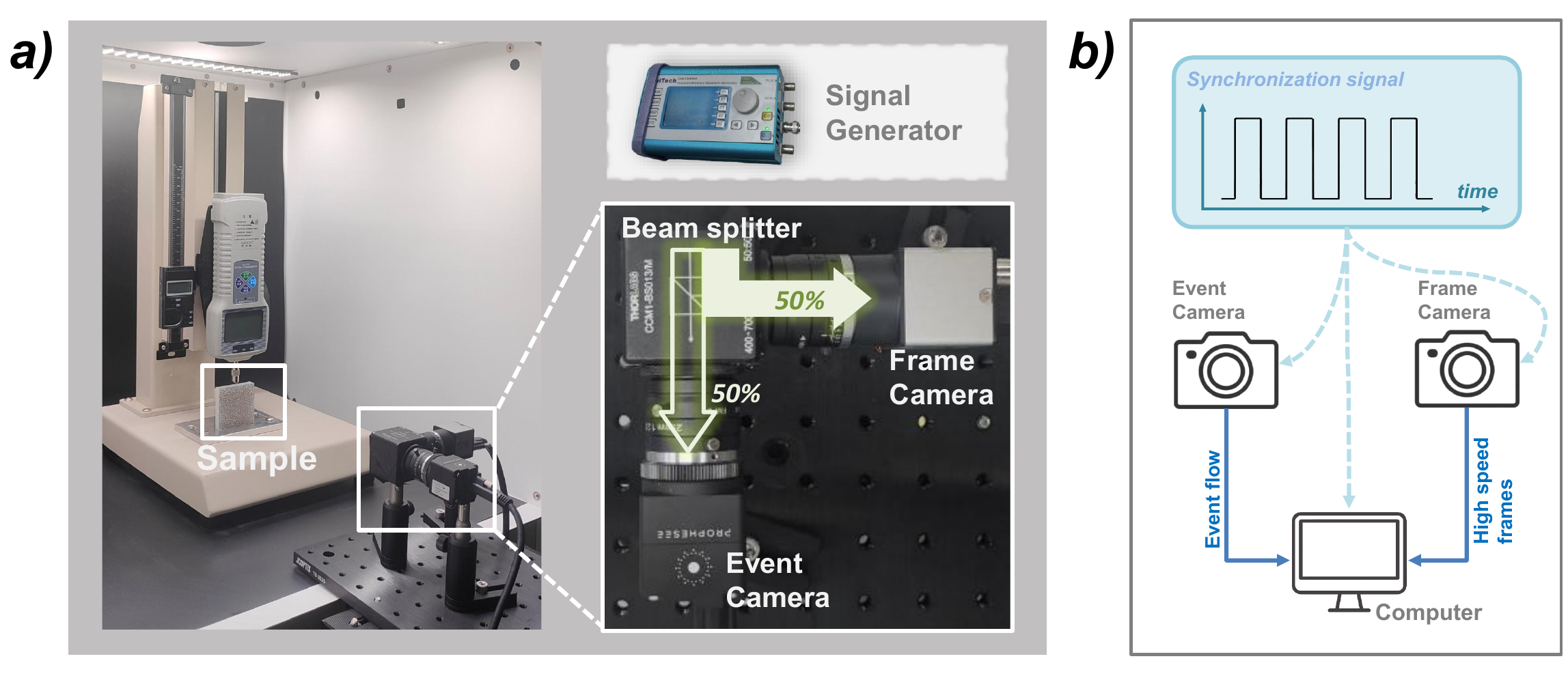}
  \caption{\textbf{Hardware configuration of the hybrid event-frame system.} \textbf{(a)} Left: Measurement setup where a pressure/stretching machine applies controlled force to the rubber sample. Right: The hybrid system comprises an event camera (Prophesee EVK4) and a 210fps grayscale camera with identical pixel size and flange distance. A signal generator produces square wave synchronization signals. The cameras are collocated by mounting a 50:50 split ratio beam splitter in front of them. Spatial calibration is performed before each data acquisition. \textbf{(b)} Data transmission topology of the system. A 210Hz square wave synchronization signal is used for triggering the frame camera and synchronizing the event stream. Simultaneously acquired event streams and frame data are transmitted to the computer for processing.}
  \label{fig:hardware}
\end{figure}

Existing real-world VDM datasets are scarce and lack corresponding event data. To comprehensively evaluate our method, we collect a new dataset using our hybrid system (Fig. \ref{fig:hardware}), which comprises temporally aligned event streams and high-frame-rate videos (210 fps). The dataset encompasses over 120 diverse deformation motions sequences spanning squeezing, stretching, bending, and cracking, covering diverse scenarios from small-scale (less than 20 pixels) to large-scale displacements (100+ pixels), as well as complex coupled deformation-egomotion dynamics. Accurate ground truth is obtained via VDM algorithms \cite{jiang2023opencorr} on high-speed frame recordings.

\begin{table*}[t!]
\centering
\resizebox{\textwidth}{!}{
\begin{tabular}{lC{1.5cm}C{1cm}C{1cm}C{1cm}C{1.5cm}C{1cm}C{1cm}C{1.5cm}C{1cm}C{1cm}C{1.5cm}}
  \toprule
  \multirow{2}*{Method} & \multicolumn{2}{c}{Data input}  & \multicolumn{3}{c}{{5-20 pixels}} & \multicolumn{3}{c}{20-100 pixels} & \multicolumn{3}{c}{100+ pixels}\\
  \cmidrule(lr){2-3}  \cmidrule(lr){4-6} \cmidrule(lr){7-9} \cmidrule(lr){10-12}
 & \small{Event}& \small{Frame}& \small{EPE} $\downarrow$ & \small{$\text{SEPE}$ $\downarrow$} & \small{Survival} $\uparrow$& \small{EPE} $\downarrow$ & \small{$\text{SEPE}$ $\downarrow$} & \small{Survival} $\uparrow$& \small{EPE} $\downarrow$ & \small{$\text{SEPE}$ $\downarrow$} & \small{Survival} $\uparrow$\\
\midrule
Opencorr \cite{jiang2023opencorr} &  & $\checkmark$  & \textbf{0.132} & \textbf{0.108} & 99.1\% & 1.727 & 0.478& 82.5\% & 39.931 & 2.782 & 25.0\%\\
Opencorr \cite{jiang2023opencorr} + Timelens \cite{tulyakov2021time} & $\checkmark$ & $\checkmark$  & 0.227 & 0.226 & \textbf{99.5\%} & 0.819 & 0.657 & 89.8\% &3.830 &1.201 & 41.3\%\\
StrainNet \cite{boukhtache2021deep} &  & $\checkmark$  & 1.553 & 1.146 & 91.0\% & 3.413& 1.230& 61.2\% & 41.870& 1.138 &18.8\%\\
StrainNet \cite{boukhtache2021deep} + Timelens \cite{tulyakov2021time} & $\checkmark$ & $\checkmark$  & 0.978& 0.911 & 95.2\% & 1.745 & 1.603 & 91.8\% & 5.189 & 2.315 & 35.4\%\\
E-Raft \cite{gehrig2021raft} & $\checkmark$&   &6.574&1.510 &21.7\% & 21.422 & 1.632 & 6.7\% & 60.312& 1.700 & 2.1\%\\
Cotrackerv3 \cite{karaev2025cotracker3} &  & $\checkmark$  & 0.671& 0.671 & 99.0\% & 2.138 & 2.026 & 91.7\% & 8.763 & 2.150 & 45.2\%\\
Cotrackerv3 \cite{karaev2025cotracker3} + Timelens \cite{tulyakov2021time}& $\checkmark$& $\checkmark$  & 0.784 & 0.783 & 99.1\% & 2.387 & 1.825 & 87.5\% & 10.788 & 2.278 &29.1\%\\
\midrule
Ours& $\checkmark$ & $\checkmark$  & 0.155& 0.121 & 99.4\% & \textbf{0.330} & \textbf{0.211} &\textbf{92.4\%} & \textbf{3.204} & \textbf{0.813} &\textbf{65.7\%}\\

\bottomrule
\end{tabular}
}
\vspace{-5pt}
\caption{\textbf{Quantitative comparisons} of our method to the model-based \cite{jiang2023opencorr} and learning-based \cite{boukhtache2021deep} VDM algorithms, event-based dense optical flow methods \cite{gehrig2021raft}, and long-term point tracking algorithms \cite{karaev2025cotracker3}. Image-based methods are enhanced with event-based frame interpolator \cite{tulyakov2021time} for thorough comparison. In terms of metrics, lower is better for EPE and SEPE, while higher is better for Survival Rate.}
\vspace{-8pt}
\label{tab:exp}
\end{table*}

\subsection{Evaluation}
\begin{figure*}[tp]
  \centering
  \includegraphics[width=0.95\linewidth]{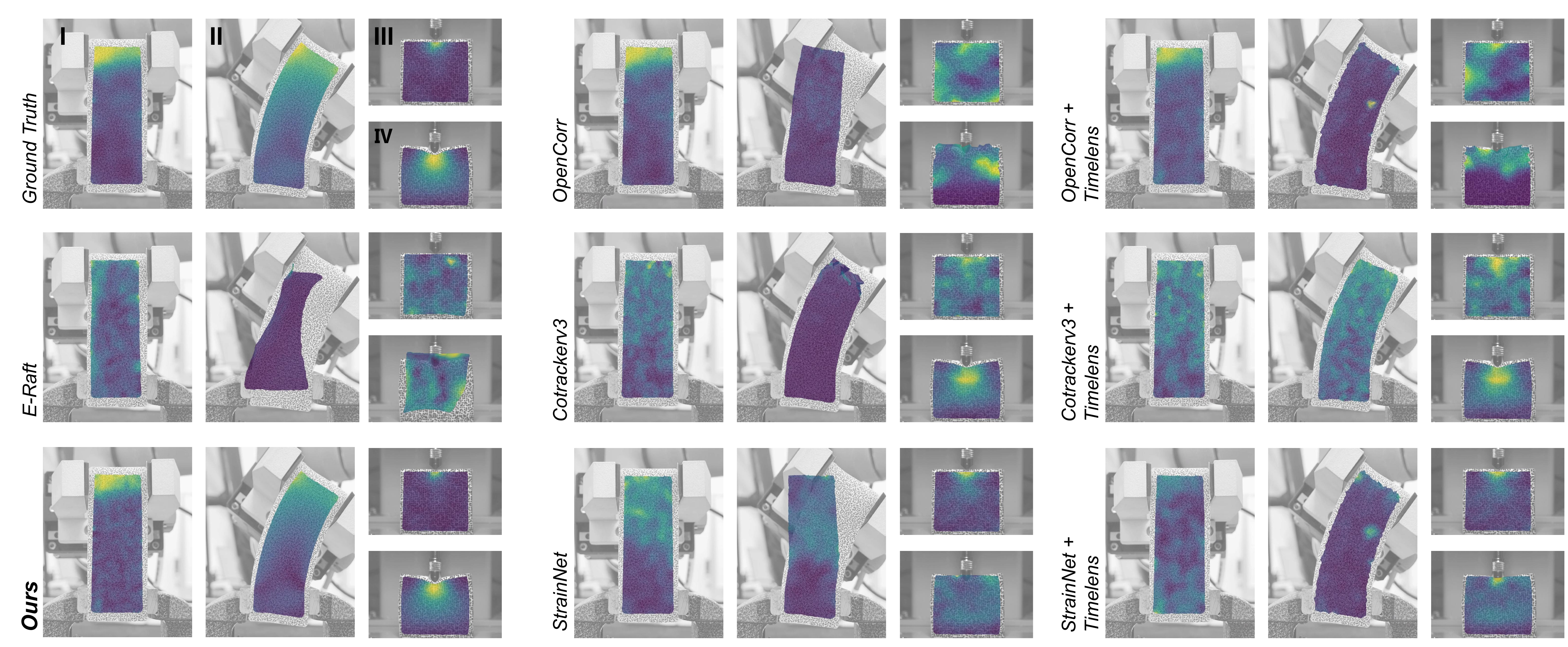}
  \caption{\textbf{Qualitative comparisons.} Results on samples under clamping and twisting (I, II), and subjected to tip pressure (III, IV).}
  \label{fig:main_exp}
  \vspace{-3mm}
\end{figure*}
\textbf{Baselines.} On the collected dataset, we benchmark our approach against several representative methods: OpenCorr~\cite{jiang2023opencorr}, a widely-used open-source model-based VDM algorithm; StrainNet~\cite{boukhtache2021deep}, a learning-based VDM method; E-RAFT~\cite{gehrig2021raft}, an event-based optical flow method for short-term motion estimation; and CoTrackerV3~\cite{karaev2025cotracker3}, a long-term point tracking method.
Given that these methods produce outputs in different formats (e.g., displacement fields, optical flow, point tracks), we adopt a unified evaluation framework to ensure fair comparison. Specifically, a dense set of points is uniformly sampled within the measurement ROI and their displacement $\{u\}$ are tracked across all frames using the respective method's output. We evaluate performance by computing metrics on the predicted trajectories against ground truth annotations.
Additionally, to ensure fairness in terms of input information, we evaluate the three image-based methods (OpenCorr~\cite{jiang2023opencorr}, StrainNet~\cite{boukhtache2021deep}, CoTracker~\cite{karaev2025cotracker3}) using high-frame-rate videos (200fps) generated by TimeLens~\cite{tulyakov2021time}, an interpolation method that synthesizes intermediate frames from the low-frame-rate videos and event streams.

\textbf{Evaluation metrics.} We report three metrics: end-point error (EPE), survival rate, and survival EPE ($\text{SEPE}$).
The EPE is computed as: $\mathrm{EPE}=\frac{1}{n} \sum_{i}\left\|u(i)-u_{g t}(i)\right\|_{2},$
which measures the average accuracy of the optical flow estimation across all points.
Following the evaluation protocol in \cite{zheng2023pointodyssey}, the survival rate measures the average time until dense tracking failure as a fraction of the entire sequence duration, where a tracking failure occurs when the L2 distance of 20\% of the sampled points exceeds 5 pixels.
Finally, $\text{SEPE}$ computes the EPE exclusively over survival points, serving as an accuracy indicator for the deformation field within the convergence region.

\textbf{Quantitative results.} Table \ref{tab:exp} presents the experimental results with event and 5 fps frame inputs. 
For small deformations (5-20 pixels), most methods maintain an EPE below $1.0$ and achieve survival rates above 95\%.
However, as the deformation magnitude increases, all methods exhibit a decline in both tracking accuracy and survival rate.
In scenarios with large continuous displacements (100+ pixels), the sota baseline, cotrackerv3, achieves a survival rate of only 45.2\%, whereas our method maintains a survival rate of 65.7\% with an EPE of 3.204 and $\text{SEPE}$ of 0.813,  demonstrating robust measurement capability under large motions and displacements.
Fig.~\ref{fig:main_exp} presents the visualization results of each parallel method. Two representative scenarios are selected: large deformation samples (100+ pixels) under clamping and twisting (I, II), and small deformation samples (5-20 pixels) subjected to tip pressure.

\subsection{Ablation Study}
\begin{figure*}[tp]
  \centering
  \includegraphics[width=0.95\linewidth]{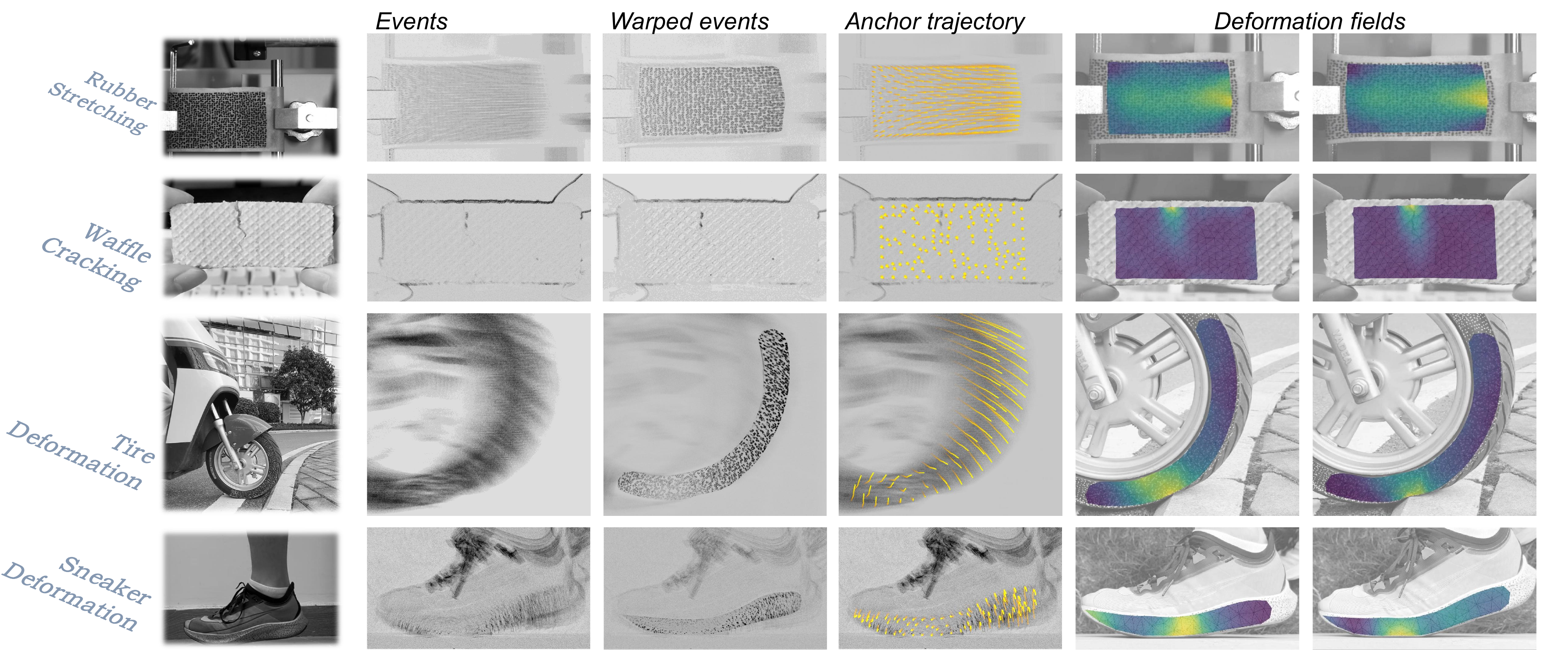}
  \caption{\textbf{Results of our system across diverse scenarios.} Each row corresponds to a different deformation scenario. From left to right, each column shows: (i) the scene overview, (ii) event count map before motion estimation, (iii) warped event count map after optimization, (iv) regressed subregion anchor trajectories, and (v) von Mises deformation fields at two time instances during the deformation process.}
  \label{fig:appl}
  \vspace{-5mm}
\end{figure*}

\textbf{Data storage burden.} We compare the data storage consumption of our method in Table \ref{tab:datasize}. Our method demonstrates significant storage efficiency, requiring only 13\% of the storage consumed by 210 fps high-speed cameras, while processing 5 fps frame and event inputs and achieving a competitive survival rate of 65.7\% and an $\text{SEPE}$ of 0.813.
\begin{table}[t!]
\centering
\resizebox{1.0\linewidth}{!}{%
\begin{tabular}{ccccccccc}
  \toprule
  \multicolumn{3}{c}{Data Input} & \multicolumn{3}{c}{{Data Size }} & \multirow{2}{*}{\small{EPE} $\downarrow$}& \multirow{2}{*}{\small{$\text{SEPE}$} $\downarrow$}& \multirow{2}{*}{Survival $\uparrow$}\\
  \cmidrule(lr){1-3}
  Event & Frame & Frame Fps & \multicolumn{3}{c}{( frames + events )} & & & \\
\midrule
$\checkmark$ & $\checkmark$ & 1 & \multicolumn{3}{c}{5.6Mb + 78.3Mb} & 4.710 & 2.533 & 32.6\% \\
\rowcolor{green!12}$\checkmark$ & $\checkmark$ & 5 & \multicolumn{3}{c}{28.1Mb + 78.3Mb} & 3.204 & 0.813 & 65.7\%\\
$\checkmark$ & $\checkmark$ & 10 & \multicolumn{3}{c}{56.3Mb + 78.3Mb} & 2.110 &0.795& 67.5\% \\
$\checkmark$ & $\checkmark$ & 20 & \multicolumn{3}{c}{112.6Mb + 78.3Mb} & 1.618 & 0.573 & 71.2\% \\
\midrule
 & $\checkmark$ & 5 & \multicolumn{3}{c}{28.1Mb} & 39.931 & 2.782 & 25.0\% \\
\rowcolor{blue!12} & $\checkmark$ & 100 & \multicolumn{3}{c}{562.5Mb} & 3.317 & 0.825 & 64.3\%\\
 & $\checkmark$ & 210 & \multicolumn{3}{c}{1181.3Mb} & \multicolumn{3}{c}{ground truth calculation}\\
\bottomrule
\end{tabular}%
}
\caption{\textbf{Comparison of data storage and processing consumption} on the 100+ pixel displacement test set. The upper panel illustrates the performance of our method with event and various frame rate inputs, while the lower shows the performance of the baseline method \cite{jiang2023opencorr} used for ground truth calculation at different frame rates. The green bars represent our method with 5 fps input, requiring only 18.9\% data storage, while the blue bars present the baseline method at 100 fps with comparable performance.}
\vspace{-10pt}
\label{tab:datasize}
\end{table}

\begin{figure}[tp]
  \centering
  \includegraphics[width=1.0\linewidth]{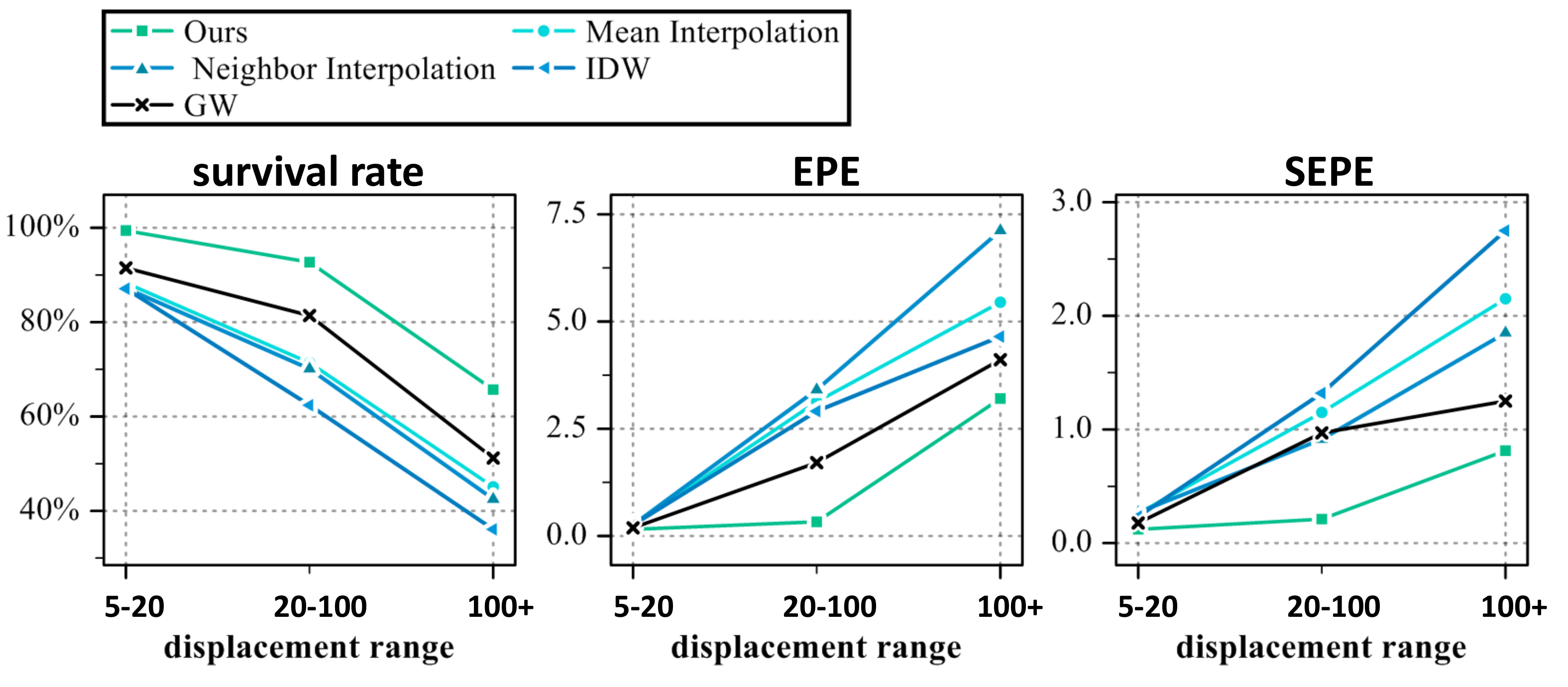}
  \caption{\textbf{Ablation study of displacement interpolation methods.} Performance comparison of our affine-invariant interpolation method against mean interpolation, neighborhood interpolation, Inverse Distance Weighting (IDW), and Gaussian Weighted Interpolation (GWI). Our method demonstrates superior measurement accuracy across different interpolation strategies.}
  \label{fig:ablation_interp}
  \vspace{-16pt}
\end{figure}

\textbf{Event displacement interpolation.} To validate the effectiveness of our affine-invariant displacement interpolation, we conduct ablation experiments comparing it against four common interpolation methods: (1) Nearest Neighbor, (2) Mean Interpolation, (3) Inverse Distance Weighting, and (4) Gaussian Weighted Interpolation. All experiments are conducted under identical settings except for the interpolation method. As shown in Fig.~\ref{fig:ablation_interp}, our method consistently achieves superior deformation measurement accuracy and survival rate compared to non-affine-invariant alternatives.

\textbf{Optimization strategies.} We compare the proposed neighborhood-greedy optimization method with the vanilla optimization method on the entire test set in terms of survival rate and mean convergence time between consecutive frames. As shown in Table \ref{tab:greedy ablation}, the neighborhood-greedy strategy effectively improves the survival rate from 49.0\% to 87.1\% by reducing error accumulation. Moreover, when incorporating convergence quality assessment for global convergence determination, it achieves a threefold speedup in mean convergence time (from 26.5s to 7.2s).
\begin{table}[t]
\resizebox{1.0\linewidth}{!}{
\setlength{\tabcolsep}{1.5 mm}
\centering
\begin{tabular}{@{}c|c|c@{}}
\toprule
Strategy & Survival $\uparrow$ & mean converge time $\downarrow$ \\ 
\midrule
Neighborhood-greedy optimization  & 65.7 \%& 7.2s\\
Vanilla optimization& 39.0 \%& 26.5s \\
\bottomrule
\end{tabular}
}
\vspace{-2mm}
\caption{\textbf{Ablation studies on optimization strategies.} Experiments are conducted using the Adam optimizer with PyTorch framework on a single NVIDIA RTX 4090 GPU (24GB).}
\label{tab:greedy ablation}
\vspace{-5mm}
\end{table}

\subsection{Measurement Results}
We evaluated our system across multiple representative scenarios to demonstrate the capabilities and application of event-based VDM (see Fig. \ref{fig:appl}), including: (a) \textbf{Rubber band stretching}, which exhibits large-scale elastic deformation that poses significant challenges to VDM methods; (b) \textbf{Wafer cracking}, featuring abrupt fracture and discontinuous motion at crack sites; (c) \textbf{Tire deformation}, relevant to vehicle safety assessment and predictive maintenance; and (d) \textbf{Sneaker deformation}, applicable to gait analysis and athletic performance evaluation. These examples demonstrate the broad applicability of our method from material testing to real-world monitoring applications.

\subsection{Limitations}
\begin{figure}[htp]
  \centering
  \includegraphics[width=1.0\linewidth]{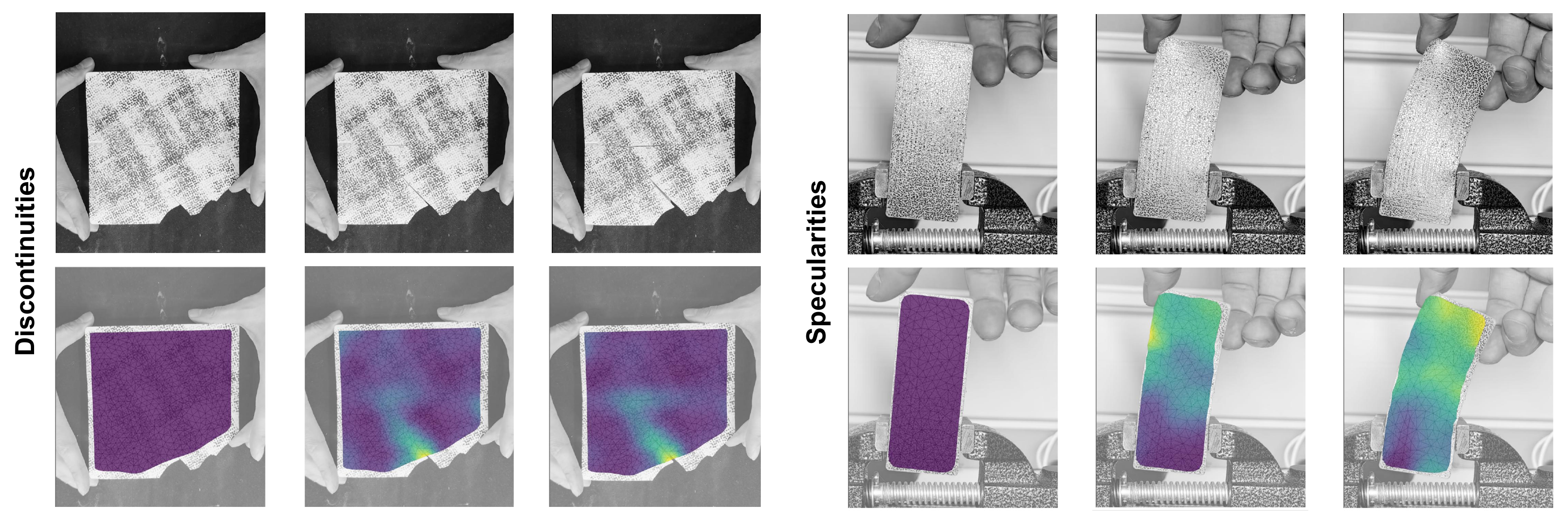}
  \caption{Failure cases of our system are demonstrated in scenarios involving plasterboard cracking and specular reflection.}
  \label{fig:failure}
  \vspace{-16pt}
\end{figure}
It is worth noting that our method has certain limitations. Relying on \textbf{spatial continuity} and \textbf{brightness constancy} assumptions, it may generate pseudo-strains when the target undergoes topological changes (e.g., cracking) and experience performance degradation in scenarios with specular reflections. We illustrate examples in Fig.~\ref{fig:failure}.

\section{Conclusion}

This paper presents a novel visual deformation measurement approach that effectively integrates events and frames. Our method demonstrates high accuracy in measuring large-scale deformations and displacements in open-world scenarios, significantly advancing the capability and applicability of visual deformation measurement.
Future work will explore several promising directions to enhance the generality and efficiency of event-based VDM. First, extending the framework to 3D deformation fields via synchronized event camera arrays would enable volumetric strain analysis. Second, developing purely event-driven VDM systems or event-driven frame sampling strategies to further minimize data throughput. Third, deploying the system to diverse real-world applications, including robotic manipulation of deformable objects where real-time feedback is crucial, structural health monitoring in civil engineering for detecting material fatigue and damage, and biomedical imaging for soft tissue tracking during surgical procedures. These extensions would enable efficient deformation measurement across multiple challenging domains.
{
    \small
    \bibliographystyle{ieeenat_fullname}
    \bibliography{main}

@article{ri2024drone,
  title={Drone-based displacement measurement of infrastructures utilizing phase information},
  author={Ri, Shien and Ye, Jiaxing and Toyama, Nobuyuki and Ogura, Norihiko},
  journal={Nature Communications},
  volume={15},
  number={1},
  pages={395},
  year={2024},
  publisher={Nature Publishing Group UK London}
}

@article{cao2021displacement,
  title={Displacement and strain mapping for osteocytes under fluid shear stress using digital holographic microscopy and digital image correlation},
  author={Cao, Runyu and Xiao, Wen and Pan, Feng and Tian, Ran and Wu, Xintong and Sun, Lianwen},
  journal={Biomedical Optics Express},
  volume={12},
  number={4},
  pages={1922--1933},
  year={2021},
  publisher={Optica Publishing Group}
}

@article{jiang2023opencorr,
  title={OpenCorr: An open source library for research and development of digital image correlation},
  author={Jiang, Zhenyu},
  journal={Optics and Lasers in Engineering},
  volume={165},
  pages={107566},
  year={2023},
  publisher={Elsevier}
}

@article{liu2024experimental,
  title={Experimental investigation on damage of concrete beam embedded with sensor using acoustic emission and digital image correlation},
  author={Liu, Chuankun and Wei, Ya},
  journal={Construction and Building Materials},
  volume={423},
  pages={135887},
  year={2024},
  publisher={Elsevier}
}

@inproceedings{gehrig2021raft,
  title={E-raft: Dense optical flow from event cameras},
  author={Gehrig, Mathias and Millh{\"a}usler, Mario and Gehrig, Daniel and Scaramuzza, Davide},
  booktitle={2021 International Conference on 3D Vision (3DV)},
  pages={197--206},
  year={2021},
  organization={IEEE}
}

@article{gehrig2024low,
  title={Low-latency automotive vision with event cameras},
  author={Gehrig, Daniel and Scaramuzza, Davide},
  journal={Nature},
  volume={629},
  number={8014},
  pages={1034--1040},
  year={2024},
  publisher={Nature Publishing Group UK London}
}

@article{he2024review,
  title={Review of research progress and development trend of digital image correlation},
  author={He, Xindang and Zhou, Run and Liu, Zheyuan and Yang, Suliang and Chen, Ke and Li, Lei},
  journal={Multidiscipline Modeling in Materials and Structures},
  volume={20},
  number={1},
  pages={81--114},
  year={2024},
  publisher={Emerald Publishing Limited}
}

@article{luo2024strain,
  title={Strain measurement at up to 3000° C based on ultraviolet-digital image correlation},
  author={Luo, YX and Dong, YL},
  journal={NDT \& E International},
  volume={146},
  pages={103155},
  year={2024},
  publisher={Elsevier}
}

@article{mguil2024various,
  title={Various experimental applications of digital image correlation method},
  author={Mguil-Touchal, S and Morestin, F and Brunei, M},
  journal={WIT Transactions on Modelling and Simulation},
  volume={17},
  year={2024},
  publisher={WIT Press}
}

@article{yoon2021digital,
  title={Digital image correlation in dental materials and related research: A review},
  author={Yoon, Sungsik and Jung, Hyung-Jo and Knowles, JC and Lee, Hae-Hyoung},
  journal={dental materials},
  volume={37},
  number={5},
  pages={758--771},
  year={2021},
  publisher={Elsevier}
}

@article{boukhtache2021deep,
  title={When deep learning meets digital image correlation},
  author={Boukhtache, Seyfeddine and Abdelouahab, Kamel and Berry, Fran{\c{c}}ois and Blaysat, Beno{\^\i}t and Grediac, Michel and Sur, Fr{\'e}d{\'e}ric},
  journal={Optics and Lasers in Engineering},
  volume={136},
  pages={106308},
  year={2021},
  publisher={Elsevier}
}

@article{boukhtache2023lightweight,
  title={A lightweight convolutional neural network as an alternative to DIC to measure in-plane displacement fields},
  author={Boukhtache, S and Abdelouahab, K and Bahou, A and Berry, F and Blaysat, B and Gr{\'e}diac, M and Sur, Fr{\'e}d{\'e}ric},
  journal={Optics and lasers in engineering},
  volume={161},
  pages={107367},
  year={2023},
  publisher={Elsevier}
}

@article{wang2023dic,
  title={DIC-Net: Upgrade the performance of traditional DIC with Hermite dataset and convolution neural network},
  author={Wang, Yin and Zhao, Jiaqing},
  journal={Optics and Lasers in Engineering},
  volume={160},
  pages={107278},
  year={2023},
  publisher={Elsevier}
}

@article{yang2022deep,
  title={Deep DIC: Deep learning-based digital image correlation for end-to-end displacement and strain measurement},
  author={Yang, Ru and Li, Yang and Zeng, Danielle and Guo, Ping},
  journal={Journal of Materials Processing Technology},
  volume={302},
  pages={117474},
  year={2022},
  publisher={Elsevier}
}

@article{yang2023r3,
  title={R3-DICnet: an end-to-end recursive residual refinement DIC network for larger deformation measurement},
  author={Yang, Jiashuai and Qian, Kemao and Wang, Lianpo},
  journal={Optics Express},
  volume={32},
  number={1},
  pages={907--921},
  year={2023},
  publisher={Optica Publishing Group}
}

@article{katz2020new,
  title={New insights on the proximal femur biomechanics using Digital Image Correlation},
  author={Katz, Yekutiel and Yosibash, Zohar},
  journal={Journal of Biomechanics},
  volume={101},
  pages={109599},
  year={2020},
  publisher={Elsevier}
}

@article{gallego2020event,
  title={Event-based vision: A survey},
  author={Gallego, Guillermo and Delbr{\"u}ck, Tobi and Orchard, Garrick and Bartolozzi, Chiara and Taba, Brian and Censi, Andrea and Leutenegger, Stefan and Davison, Andrew J and Conradt, J{\"o}rg and Daniilidis, Kostas and others},
  journal={IEEE transactions on pattern analysis and machine intelligence},
  volume={44},
  number={1},
  pages={154--180},
  year={2020},
  publisher={IEEE}
}

@article{brandli2014240,
  title={A 240$\times$ 180 130 db 3 $\mu$s latency global shutter spatiotemporal vision sensor},
  author={Brandli, Christian and Berner, Raphael and Yang, Minhao and Liu, Shih-Chii and Delbruck, Tobi},
  journal={IEEE Journal of Solid-State Circuits},
  volume={49},
  number={10},
  pages={2333--2341},
  year={2014},
  publisher={IEEE}
}

@article{serrano2013128,
  title={A 128$\times$ 128 1.5\% Contrast Sensitivity 0.9\% FPN 3 $\mu$s Latency 4 mW Asynchronous Frame-Free Dynamic Vision Sensor Using Transimpedance Preamplifiers},
  author={Serrano-Gotarredona, Teresa and Linares-Barranco, Bernab{\'e}},
  journal={IEEE Journal of Solid-State Circuits},
  volume={48},
  number={3},
  pages={827--838},
  year={2013},
  publisher={IEEE}
}

@inproceedings{tulyakov2021time,
  title={Time lens: Event-based video frame interpolation},
  author={Tulyakov, Stepan and Gehrig, Daniel and Georgoulis, Stamatios and Erbach, Julius and Gehrig, Mathias and Li, Yuanyou and Scaramuzza, Davide},
  booktitle={Proceedings of the IEEE/CVF conference on computer vision and pattern recognition},
  pages={16155--16164},
  year={2021}
}

@inproceedings{tulyakov2022time,
  title={Time lens++: Event-based frame interpolation with parametric non-linear flow and multi-scale fusion},
  author={Tulyakov, Stepan and Bochicchio, Alfredo and Gehrig, Daniel and Georgoulis, Stamatios and Li, Yuanyou and Scaramuzza, Davide},
  booktitle={Proceedings of the IEEE/CVF Conference on Computer Vision and Pattern Recognition},
  pages={17755--17764},
  year={2022}
}

@article{liao2024ef,
  title={Ef-3dgs: Event-aided free-trajectory 3d gaussian splatting},
  author={Liao, Bohao and Zhai, Wei and Wan, Zengyu and Cheng, Zhixin and Yang, Wenfei and Zhang, Tianzhu and Cao, Yang and Zha, Zheng-Jun},
  journal={arXiv preprint arXiv:2410.15392},
  year={2024}
}

@article{wu2024event,
  title={Event-based asynchronous HDR imaging by temporal incident light modulation},
  author={Wu, Yuliang and Tan, Ganchao and Chen, Jinze and Zhai, Wei and Cao, Yang and Zha, Zheng-Jun},
  journal={Optics Express},
  volume={32},
  number={11},
  pages={18527--18538},
  year={2024},
  publisher={Optica Publishing Group}
}

@article{wan2024event,
  title={Event-based optical flow via transforming into motion-dependent view},
  author={Wan, Zengyu and Tan, Ganchao and Wang, Yang and Zhai, Wei and Cao, Yang and Zha, Zheng-Jun},
  journal={IEEE Transactions on Image Processing},
  volume={33},
  pages={5327--5339},
  year={2024},
  publisher={IEEE}
}

@article{han2024event,
  title={Event-based tracking any point with motion-augmented temporal consistency},
  author={Han, Han and Zhai, Wei and Cao, Yang and Li, Bin and Zha, Zheng-jun},
  journal={arXiv preprint arXiv:2412.01300},
  year={2024}
}

@article{agudo2015sequential,
  title={Sequential non-rigid structure from motion using physical priors},
  author={Agudo, Antonio and Moreno-Noguer, Francesc and Calvo, Bego{\~n}a and Montiel, Jos{\'e} Mar{\'\i}a Mart{\'\i}nez},
  journal={IEEE transactions on pattern analysis and machine intelligence},
  volume={38},
  number={5},
  pages={979--994},
  year={2015},
  publisher={IEEE}
}

@inproceedings{kairanda2022f,
  title={f-sft: Shape-from-template with a physics-based deformation model},
  author={Kairanda, Navami and Tretschk, Edith and Elgharib, Mohamed and Theobalt, Christian and Golyanik, Vladislav},
  booktitle={Proceedings of the IEEE/CVF Conference on Computer Vision and Pattern Recognition},
  pages={3948--3958},
  year={2022}
}

@article{xue2024event,
  title={Event-based non-rigid reconstruction of low-rank parametrized deformations from contours},
  author={Xue, Yuxuan and Li, Haolong and Leutenegger, Stefan and St{\"u}ckler, J{\"o}rg},
  journal={International Journal of Computer Vision},
  volume={132},
  number={8},
  pages={2943--2961},
  year={2024},
  publisher={Springer}
}

@inproceedings{nehvi2021differentiable,
  title={Differentiable event stream simulator for non-rigid 3d tracking},
  author={Nehvi, Jalees and Golyanik, Vladislav and Mueller, Franziska and Seidel, Hans-Peter and Elgharib, Mohamed and Theobalt, Christian},
  booktitle={Proceedings of the IEEE/CVF Conference on Computer Vision and Pattern Recognition},
  pages={1302--1311},
  year={2021}
}

@article{xue2022event,
  title={Event-based non-rigid reconstruction from contours},
  author={Xue, Yuxuan and Li, Haolong and Leutenegger, Stefan and Stueckler, Joerg},
  journal={arXiv preprint arXiv:2210.06270},
  year={2022}
}

@inproceedings{millerdurai20243d,
  title={3d pose estimation of two interacting hands from a monocular event camera},
  author={Millerdurai, Christen and Luvizon, Diogo and Rudnev, Viktor and Jonas, Andr{\'e} and Wang, Jiayi and Theobalt, Christian and Golyanik, Vladislav},
  booktitle={2024 International Conference on 3D Vision (3DV)},
  pages={291--301},
  year={2024},
  organization={IEEE}
}

@inproceedings{malti2017elastic,
  title={Elastic shape-from-template with spatially sparse deforming forces},
  author={Malti, Abed and Herzet, C{\'e}dric},
  booktitle={Proceedings of the IEEE conference on computer vision and pattern recognition},
  pages={3337--3345},
  year={2017}
}

@inproceedings{agudo2014good,
  title={Good vibrations: A modal analysis approach for sequential non-rigid structure from motion},
  author={Agudo, Antonio and Agapito, Lourdes and Calvo, Begona and Montiel, Jose MM},
  booktitle={Proceedings of the IEEE Conference on computer vision and pattern recognition},
  pages={1558--1565},
  year={2014}
}

@article{wan2025emotive,
  title={EMoTive: Event-guided Trajectory Modeling for 3D Motion Estimation},
  author={Wan, Zengyu and Zhai, Wei and Cao, Yang and Zha, Zhengjun},
  journal={arXiv preprint arXiv:2503.11371},
  year={2025}
}

@inproceedings{yang2023learning,
  title={Learning event guided high dynamic range video reconstruction},
  author={Yang, Yixin and Han, Jin and Liang, Jinxiu and Sato, Imari and Shi, Boxin},
  booktitle={Proceedings of the IEEE/CVF Conference on Computer Vision and Pattern Recognition},
  pages={13924--13934},
  year={2023}
}

@article{mostafavi2021learning,
  title={Learning to reconstruct hdr images from events, with applications to depth and flow prediction},
  author={Mostafavi, Mohammad and Wang, Lin and Yoon, Kuk-Jin},
  journal={International Journal of Computer Vision},
  volume={129},
  number={4},
  pages={900--920},
  year={2021},
  publisher={Springer}
}

@article{zou2024eventhdr,
  title={EventHDR: From Event to High-Speed HDR Videos and Beyond},
  author={Zou, Yunhao and Fu, Ying and Takatani, Tsuyoshi and Zheng, Yinqiang},
  journal={IEEE Transactions on Pattern Analysis and Machine Intelligence},
  year={2024},
  publisher={IEEE}
}

@inproceedings{hamann2024low,
  title={Low-power Continuous Remote Behavioral Localization with Event Cameras},
  author={Hamann, Friedhelm and Ghosh, Suman and Martinez, Ignacio Juarez and Hart, Tom and Kacelnik, Alex and Gallego, Guillermo},
  booktitle={Proceedings of the IEEE/CVF Conference on Computer Vision and Pattern Recognition},
  pages={18612--18621},
  year={2024}
}

@article{chen2024event,
  title={Event-Based Motion Magnification},
  author={Chen, Yutian and Guo, Shi and Yu, Fangzheng and Zhang, Feng and Gu, Jinwei and Xue, Tianfan},
  journal={arXiv preprint arXiv:2402.11957},
  year={2024}
}

@article{palanca2016use,
  title={The use of digital image correlation in the biomechanical area: a review},
  author={Palanca, Marco and Tozzi, Gianluca and Cristofolini, Luca},
  journal={International biomechanics},
  volume={3},
  number={1},
  pages={1--21},
  year={2016},
  publisher={Taylor \& Francis}
}

@article{genovese2013improved,
  title={An improved panoramic digital image correlation method for vascular strain analysis and material characterization},
  author={Genovese, Katia and Lee, YU and Lee, AY and Humphrey, JD},
  journal={Journal of the mechanical behavior of biomedical materials},
  volume={27},
  pages={132--142},
  year={2013},
  publisher={Elsevier}
}

@article{krehbiel2010digital,
  title={Digital image correlation for improved detection of basal cell carcinoma},
  author={Krehbiel, Joel David and Lambros, John and Viator, JA and Sottos, Nancy R},
  journal={Experimental Mechanics},
  volume={50},
  pages={813--824},
  year={2010},
  publisher={Springer}
}

@article{huang2010high,
  title={High-efficiency cell--substrate displacement acquisition via digital image correlation method using basis functions},
  author={Huang, Jianyong and Pan, Xiaochang and Peng, Xiaoling and Zhu, Tao and Qin, Lei and Xiong, Chunyang and Fang, Jing},
  journal={Optics and lasers in engineering},
  volume={48},
  number={11},
  pages={1058--1066},
  year={2010},
  publisher={Elsevier}
}

@article{verbruggen2015altered,
  title={Altered mechanical environment of bone cells in an animal model of short-and long-term osteoporosis},
  author={Verbruggen, Stefaan W and Mc Garrigle, Myles J and Haugh, Matthew G and Voisin, Muriel C and McNamara, Laoise M},
  journal={Biophysical journal},
  volume={108},
  number={7},
  pages={1587--1598},
  year={2015},
  publisher={Elsevier}
}

@article{zhou2016using,
  title={Using digital image correlation to characterize local strains on vascular tissue specimens},
  author={Zhou, Boran and Ravindran, Suraj and Ferdous, Jahid and Kidane, Addis and Sutton, Michael A and Shazly, Tarek},
  journal={Journal of visualized experiments: JoVE},
  volume={107}, 
  year={2016},
  publisher={MyJoVE Corporation}
}

@inproceedings{gallego2018unifying,
  title={A unifying contrast maximization framework for event cameras, with applications to motion, depth, and optical flow estimation},
  author={Gallego, Guillermo and Rebecq, Henri and Scaramuzza, Davide},
  booktitle={Proceedings of the IEEE conference on computer vision and pattern recognition},
  pages={3867--3876},
  year={2018}
}

@inproceedings{stoffregen2019event,
  title={Event cameras, contrast maximization and reward functions: An analysis},
  author={Stoffregen, Timo and Kleeman, Lindsay},
  booktitle={Proceedings of the IEEE/CVF Conference on Computer Vision and Pattern Recognition},
  pages={12300--12308},
  year={2019}
}

@article{gehrig2020eklt,
  title={EKLT: Asynchronous photometric feature tracking using events and frames},
  author={Gehrig, Daniel and Rebecq, Henri and Gallego, Guillermo and Scaramuzza, Davide},
  journal={International Journal of Computer Vision},
  volume={128},
  number={3},
  pages={601--618},
  year={2020},
  publisher={Springer}
}

@article{zhu2018ev,
  title={EV-FlowNet: Self-supervised optical flow estimation for event-based cameras},
  author={Zhu, Alex Zihao and Yuan, Liangzhe and Chaney, Kenneth and Daniilidis, Kostas},
  journal={arXiv preprint arXiv:1802.06898},
  year={2018}
}

@article{gehrig2024dense,
  title={Dense continuous-time optical flow from event cameras},
  author={Gehrig, Mathias and Muglikar, Manasi and Scaramuzza, Davide},
  journal={IEEE Transactions on Pattern Analysis and Machine Intelligence},
  volume={46},
  number={7},
  pages={4736--4746},
  year={2024},
  publisher={IEEE}
}

@article{shiba2024secrets,
  title={Secrets of event-based optical flow, depth and ego-motion estimation by contrast maximization},
  author={Shiba, Shintaro and Klose, Yannick and Aoki, Yoshimitsu and Gallego, Guillermo},
  journal={IEEE Transactions on Pattern Analysis and Machine Intelligence},
  volume={46},
  number={12},
  pages={7742--7759},
  year={2024},
  publisher={IEEE}
}

@inproceedings{hamann2024motion,
  title={Motion-prior contrast maximization for dense continuous-time motion estimation},
  author={Hamann, Friedhelm and Wang, Ziyun and Asmanis, Ioannis and Chaney, Kenneth and Gallego, Guillermo and Daniilidis, Kostas},
  booktitle={European Conference on Computer Vision},
  pages={18--37},
  year={2024},
  organization={Springer}
}

@article{hagenaars2021self,
  title={Self-supervised learning of event-based optical flow with spiking neural networks},
  author={Hagenaars, Jesse and Paredes-Vall{\'e}s, Federico and De Croon, Guido},
  journal={Advances in Neural Information Processing Systems},
  volume={34},
  pages={7167--7179},
  year={2021}
}

@inproceedings{paredes2023taming,
  title={Taming contrast maximization for learning sequential, low-latency, event-based optical flow},
  author={Paredes-Vall{\'e}s, Federico and Scheper, Kirk YW and De Wagter, Christophe and De Croon, Guido CHE},
  booktitle={Proceedings of the IEEE/CVF international conference on computer vision},
  pages={9695--9705},
  year={2023}
}

@inproceedings{karaev2025cotracker3,
  title={Cotracker3: Simpler and better point tracking by pseudo-labelling real videos},
  author={Karaev, Nikita and Makarov, Yuri and Wang, Jianyuan and Neverova, Natalia and Vedaldi, Andrea and Rupprecht, Christian},
  booktitle={Proceedings of the IEEE/CVF International Conference on Computer Vision},
  pages={6013--6022},
  year={2025}
}

@inproceedings{zheng2023pointodyssey,
  title={Pointodyssey: A large-scale synthetic dataset for long-term point tracking},
  author={Zheng, Yang and Harley, Adam W and Shen, Bokui and Wetzstein, Gordon and Guibas, Leonidas J},
  booktitle={Proceedings of the IEEE/CVF International Conference on Computer Vision},
  pages={19855--19865},
  year={2023}
}
}


\end{document}